\pdfoutput=1
\documentclass{article}
\usepackage{ijcai22}

\usepackage{times}
\usepackage{soul}
\usepackage{url}
\usepackage[utf8]{inputenc}
\usepackage[small]{caption}
\usepackage{graphicx}
\usepackage{amsmath}
\usepackage{amsthm}
\usepackage{amssymb}
\usepackage{booktabs}
\usepackage{algorithm}
\usepackage{algorithmic}
\usepackage{bm}
\usepackage{multirow}
\usepackage{booktabs}
\usepackage[T1]{fontenc}
\usepackage[dvipsnames]{xcolor}
\usepackage{lipsum}
\usepackage{xcolor}         
\usepackage{caption}
\usepackage{enumitem}
\usepackage{pifont}
\newcommand{\cmark}{\ding{51}}

\title{Regularized Graph Structure Learning with Semantic Knowledge \\
	 for Multi-variates Time-Series Forecasting}

\author{Hongyuan Yu$^{123}$\footnote{equal contribution, This work was partly done when Hongyuan Yu was intern at Ant Group.}, Ting Li$^{1*}$, Weichen Yu$^{23*}$, Jianguo Li$^{1}$\footnote{corresponding author}, Yan Huang$^{23}$, Liang Wang$^{23\dag}$, Alex Liu$^{1}$
\affiliations
$^1$ Ant Group \\
$^2$ AI School, University of Chinese Academy of Sciences \\
$^3$ CRIPAC, NLPR, Institute of Automation, Chinese Academy of Sciences, China
\emails
\{hongyuan.yu,weichen.yu\}@cripac.ia.ac.cn,
\{lt317068, lijg.zero, alexliu\}@antgroup.com,  \\
\{yhuang, wangliang\}@nlpr.ia.ac.cn
}

\begin{document}

\maketitle

\begin{abstract}
  Multivariate time-series forecasting is a critical task for many applications, and graph time-series network is widely studied due to its capability to capture the spatial-temporal correlation simultaneously.
  However, most existing works focus more on learning with the explicit prior graph structure, while ignoring potential information from the implicit graph structure, yielding incomplete structure modeling. Some recent works attempts to learn the intrinsic or implicit graph structure directly, while lacking a way to combine explicit prior structure with implicit structure together. 
   In this paper, we propose Regularized Graph Structure Learning (RGSL) model to incorporate both explicit prior structure and implicit structure together, and learn the forecasting deep networks along with the graph structure. 
   RGSL consists of two innovative modules. First, we derive an implicit dense similarity matrix through node embedding, and learn the sparse graph structure using the Regularized Graph Generation (RGG) based on the Gumbel Softmax trick. 
   Second, we propose a Laplacian Matrix Mixed-up Module (LM$^3$) to fuse the explicit graph and implicit graph together. 
We conduct experiments on three real-word datasets. Results show that the proposed RGSL model outperforms existing graph forecasting algorithms with a notable margin, while learning meaningful graph structure simultaneously. Our code and models are made publicly available at \url{https://github.com/alipay/RGSL.git}. 
\end{abstract}

\section{Introduction}
The spatial-temporal graph network~\cite{stgcn,cao2020spectral,chen2019gated,shang2021discrete} enhances time-series forecasting by modelling the correlation as well as the relationship between multivariate  time-series. 
It has many applications. For example,  the traffic flow forecasting is a basic yet important application in intelligent transportation system, which constructs the spatial dependency graph with road distance. The cloud service flow forecasting is a fundamental task in cloud serving system and e-commerce domain, which builds the relationship graph based on request region or zone information. %
Existing graph time-series forecasting networks like STGCN~\cite{stgcn}, DCRNN~\cite{DCRNN} and ASTGCN~\cite{guo2019attention} exploit fixed graph structure constructed with domain expert knowledge to capture the multi-variate time-series relationship. 
The explicit graph is not always available in every applications or may be incomplete as it is hard for human expert to capture latent or long-range dependence among substantial time-series. 
Thus how to define accurate dynamic relationship graph becomes a critical task for graph time-series forecasting. 

 \begin{figure}[!t]
\centering
\includegraphics[width=1.\linewidth]{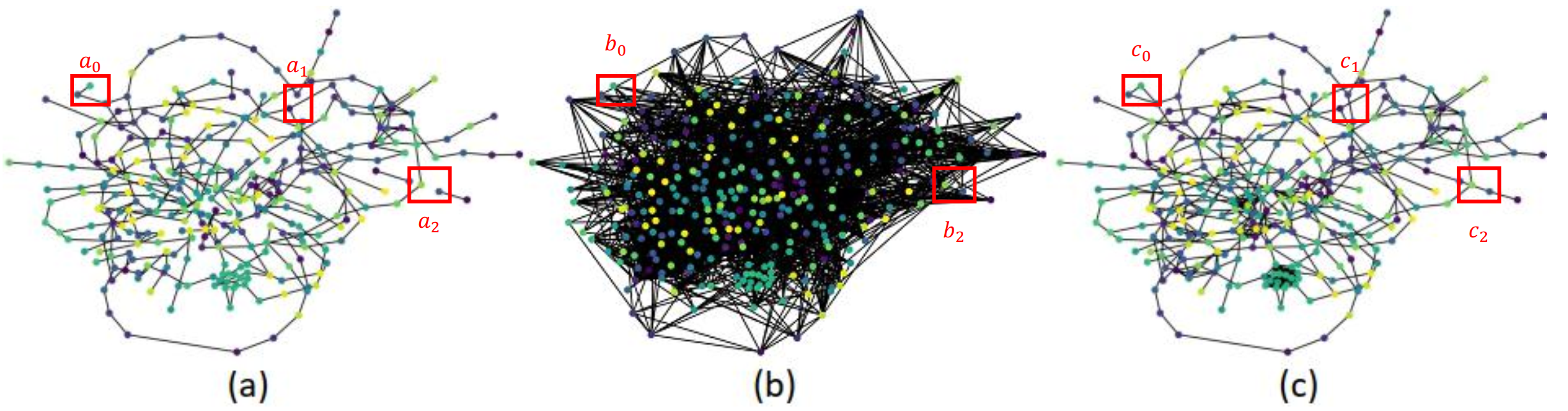}
\caption{Graph Visualization (a) naive explicit GSL trained from prior knowledge; (b) implicit GSL from popular network AGCRN; (c) our proposed RGSL}
\label{fig:visual}
\vskip -0.2in
\end{figure}

As the quality of graph structure impacts the performances of graph time-series forecasting greatly, many recent efforts      ~\cite{dstgcn2021,agcrn,chen2021gog} have made for Graph Structure Learning (GSL).
For instance, GTS ~\cite{shang2021discrete} have been proposed to learn the discrete graph structure simultaneously with GNN. 
AGCRN in ~\cite{agcrn} is proposed to learn the similarity matrix derived from trainable adaptive node embedding and forecasting in an end-to-end style. DGNN ~\cite{dstgcn2021} is  a dynamic graph construction method  which  learns the time-specific spatial adjacency matrix firstly and then exploits dynamic graph convolution to pass the message.
However, these aforementioned methods go to another extreme that they learn the intrinsic/implicit graph structure from time-series patterns directly, while ignoring the possibility to leverage priori time-series relationships defined from domain expert knowledge.

In this paper, we focus on solving two problems, the first is how to take advantage of combining the explicit time-series relationship with implicit correlations effectively in an end-to-end way; the second is how to regularize the learned graph to be sparse which filters out the redundant useless edges thus improves overall performances, and is more valuable to real-world applications. 
To address these issues, firstly we introduce the Regularized Graph Generation (RGG) module to learn the implicit graph, which adopts the Gumbel Softmax trick to sparsify the dense similarity matrix from node embedding. 
Second, we introduce the Laplacian Matrix Mixed-up Module (LM$^3$) to incorporate the explicit relationship from domain knowledge with the implicit graph from RGG.  
Figure\ref{fig:visual} shows the graph structure learned from only explicit relationship in (a), both implicit and explicit relationship without regularization (b), as well as from our proposed RGSL shown in (c). 
We can observe that RGSL can discover the implicit time-series relationship ignored by naive graph structure learning algorithm(shown in red boxes in Figure\ref{fig:visual}(a)).
Besides, compared to Figure \ref{fig:visual}(b), the regularization module in RGSL which automatically removes the noisy/redundant edges making the learned graph more sparse, as well as more effective than dense graph.

To summarize, our work presents the following contributions.
 \begin{itemize}
 \item We propose a novel and efficient model named RGSL which first exploits both explicit and implicit time-series relationship to assist graph structure learning, and our proposed LM$^3$ module effectively mixes up two kinds of Laplacian matrix collectively.
 \item Besides, to regularize the learned matrix, we also propose a RGG module which formulates the discrete graph structure as a variable independent matrix and exploits the Gumbel softmax trick to optimize the probabilistic graph distribution parameters.
 \item Extensive experiments show the proposed model RGSL significantly outperforms benchmarks  on three datasets consistently. Moreover, both the LM$^3$ module and RGG module can be easily generalized to different spatio-temporal graph models.
 \end{itemize}
\vspace{-0.1cm}
\section{Methodology}
\begin{figure*}[t]
		\centering
		\includegraphics[scale=0.68]{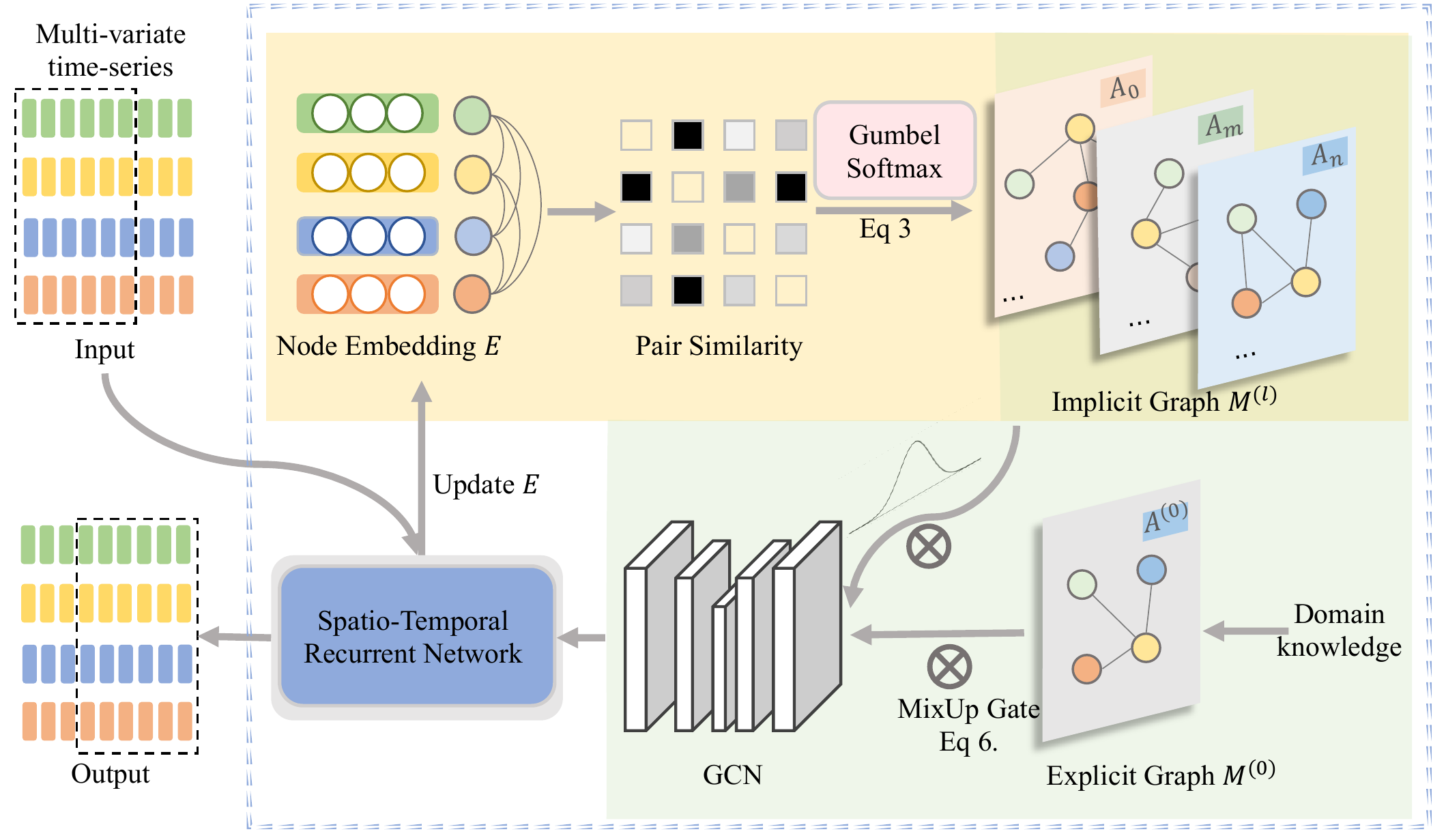}
		\caption{The framework of RGSL: it consists of three modules. Regularized Graph Generation(RGG) module (inside the yellow box) takes trainable node embedding $\bm{E}$ as input and outputs learned discrete graph $\bm{A^{(l)}}$; then Laplacian Matrix Mix-up Module(LM$^3$) (inside the green box) outputs adjacent matrix $\bm{A}$ which encodes both explicit knowledge $\bm{M^{(0)}}$ and implicit knowledge $\bm{M^{(l)}}$. $\bm{A}$ goes through spatial-temporal graph convolution recurrent network(inside the blue dashed box) to give the forecasting results.}
		\label{fig:model}
		\vspace{-0.3cm} 
	\end{figure*}

In this section, we first introduce problem definition and notations, and then describe the detailed implementation of proposed RGSL. The overall pipeline is shown in Figure \ref{fig:model}.
The RGSL consists of three major modules, and the first is the regularized graph generation module name RGG which learns the discrete graph structure with trainable node embeddings in Section \ref{sec:Graph Generation} with Gumbel softmax trick. 
The second is the Laplacian matrix mix-up module named LM$^3$ in \ref{sec:MAM} which captures both explicit and implicit time-series correlations between nodes in a convex way. 
Finally, in \ref{section:spa-tem}, we utilize recurrent graph network to perform time-series forecasting considering both the spatial correlation and temporal dependency simultaneously.

\vspace{-0.1cm}
\subsection{Preliminary}
The traffic series forecasting is to predict the future time series from historical traffic records. Denote the training data by $\bm{{X_{0:T}}} = \{\bm{X_{0}}, \bm{X_{1}},\ldots,\bm{X_{t}},\ldots, \bm{X_{T}}\}$, and $\bm{X_{t}} = \{\bm{X_{t}^{0}}, \bm{X_{t}^{1}},\ldots, \bm{X_{t}^{N}}\}$, where the superscript refers to series and subscript refers to time. There are total T timestamps for training and $\bm{\tau}$ timestamps required for traffic forecasting. 
We denote $\bm{\mathcal{G}^{(0)}}$ as the explicit graph constructed with priori time-series relationship, and $\bm{\mathcal{G}^{(l)}}$ as the implicit graph learned from trainable node embeddings, and the vertex of graph $\bm{\mathcal{G}^{(l)}}$ represents traffic series $\bm{\mathcal{X}}$, and $\bm{A} \in \mathbb{R}^{N \times N}$ is the adjacent matrix of the graph $\bm{\mathcal{G}}$ representing the similarity between time-series. Thus, the time-series forecasting with the explicit graph task can be defined as:
\begin{equation}
\small
\min_{W_\theta} \bm{\mathcal{L}}(\bm{X}_{T+1:T+\tau},\bm{\hat{X}}_{T+1:T+\tau};
\bm{X}_{0:T},\bm{\mathcal{G}}^{(0)}, \bm{\mathcal{G}}^{(l)})
\end{equation}
where ${W_\theta}$ denotes all the learnable parameters, $\bm{\hat{X}}_{T+1:T+\tau}$ denotes the ground truth future values,  $\bm{\mathcal{L}}$ is the loss function.

\subsection{Regularized Graph Generation}\label{sec:Graph Generation}

Regularization method Dropout~\cite{dropout} aims at preventing neural networks from overfitting by randomly drop connections during training. However, traditional Dropout equally treats every connection and drop them with the same distribution acquired from cross-validation, which doesn't consider the different significance of different edges. In our Regularized Graph Generation(RGG) module, inspired by~\cite{shang2021discrete} and works in reinforcement learning, we simply resolve the regularization problem with Gumble Softmax to replace Softmax, which is super convenient to employ, increases the explainability of prediction and shows nice improvements. Another motivation of applying Gumble Softmax trick is to alleviate the density of the learned matrix after training from GNNs.

Let $\bm{E} \in \mathbb{R}^{N \times d}$ be the learned node embedding matrix, $d$ is the embedding dimension, $\bm{\theta}$ is the probability matrix then $\bm{\theta}_{ij} \in \bm{\theta}$ represents the probability to preserve the edge of time-series $i$ to $j$, which is formulated as:
\begin{equation} \label{equ:nodes}
\small
 \bm{\theta} = \bm{E}\bm{E}^\top
\end{equation}
Let $\sigma$ be activation function and $s$ is the temperature variable, then the sparse adjacency matrix $\bm{A}^{(l)}$ is defined as:
\begin{equation} \label{equ:gumbel}
\small
\begin{aligned}
\bm{A}^{(l)} = \ & \sigma((\log(\bm{\theta}_{ij} / (1-\bm{\theta}_{ij})) + (g_{ij}^{1} - g_{ij}^{2})) / s) \\
& s.t. \quad g_{ij}^{1}, g_{ij}^{2} \sim Gumbel(0, 1)
\end{aligned}
\end{equation}
Equation \ref{equ:gumbel} is the Gumbel softmax implementation of our task where the $\bm{A}^{(l)}_{ij}=1$ with the probability $\bm{\theta}_{ij}$ and 0 with remaining probability.
It can be easily proved that Gumbel Softmax shares the same probability distribution as the normal Softmax, which ensures that the graph forecasting network keeps consistent with the trainable probability matrix generation statistically. 

At each iteration, we calculate the adjacent matrix $\bm{\theta}$ as the Equation \ref{equ:nodes} suggests, Gumbel-Max samples the adjacent matrix to determine which edge to preserve and which to discard, which is similar to Dropout. However, dropout randomly selects edges or neurons with equal probability, while we drop out useful edges with small likelihood and tend to get rid of those redundant edges. As in Figure. \ref{fig:adjsparce}(a), all the non-diagonal entries are non-zero, but substantial amounts of them are small-value and regarded as useless or even noisy. 
Another difference from dropout is that in test phase, RGG also utilizes Gumbel Softmax to remove the noise information contained in redundant small values. 
In this way, we filter out the similarity information between nodes which is beneficial for later traffic predicting, and also inherit the advantages of dropout such as improving the regularization and generalization, and preventing too much co-adapting. 
From another perspective, RGG also improves the explainability in graph time-series prediction since $\bm{A}^{(l)}$ is sparse which is more similar to semantic knowledge shown in Figure. \ref{fig:visual}. 

The RGG module regularizes the learnable graph generated with trainable node embeddings, making it more explainable. It is easy to add to graph generation tasks, significantly reducing the computation cost.

\vspace{-0.1cm}
\subsection{Laplacian Matrix Mix-up Module} \label{sec:MAM}
There are many kinds of attributes in time-series, for example in traffic time-series forecasting task, each node $\bm{X_{0:T}^{i}}$ represents flows in a POI, whose properties are affected by multiple complicated factors: location, time of the day, nearby commercial buildings, and so on. We classify the factors into three categories: 
1) temporal dependencies; 2) spatial dependencies; 3) external factors. 
The first two factors are easy to understand, and external factors can be illustrated in this example: although road A and road B are not connected spatially, they are both six-lane avenues and have gas stations on them, thus we assume A and B may share similar patterns. 
We propose a Laplacian matrix mix-up module named LM$^3$ which carefully fuses both the explicit graph generated from priori relationship and the implicit graph calculated with trainable node embeddings, discovering the above three categories of relationship.

The spatial adjacent information is often provided with datasets themselves, or can be acquired with little effort from urban maps. 
Let's denote them as explicit graph $\bm{A}^{(0)}$, then the spatially adjacent relationship is easily captured by$\{{\mathbf{X}_{0:T}}^{j}\}, \ s.t.\ \bm{A}_{i,j}^{(0)}=1$. Although the explicit graph may not include complete information, it contains spatial correlations and can be set as the training start-point. 
On the other hand, the $\bm{A}^{(l)}$, derived from Eq.\ref{equ:gumbel}, encodes the inner product of node embeddings, and captures the implicit correlation between multivariate time-series. 
Let $\bm{X} \in \mathbb{R}^{N \times H}$ be input matrix, $\bm{W}^{(0)}$, $\bm{W}_{b}^{(0)}$, $\bm{W}^{(l)}$, $ \bm{W}_{b}^{(l)}$ are learnable parameters, then the Chebyshev polynomial expansion form of graph operations for $\bm{A}^{(l)}$ (Eq.\ref{equ:learn}) and $\bm{A}^{(0)}$ (Eq.\ref{equ:pre})
\begin{equation}\label{equ:pre}
    \bm{\Tilde{A}}^{(0)}(\bm{X}) = (\bm{I} + \bm{D}^{-\frac{1}{2}} \bm{A}^{(0)} \bm{D}^{-\frac{1}{2}}) \bm{X} \bm{W}^{(0)} + \bm{W}_{b}^{(0)}
\end{equation}
\begin{equation}\label{equ:learn}
    \bm{\Tilde{A}}^{(l)}(\bm{X}) =  (\bm{I} + \bm{D}^{-\frac{1}{2}} \bm{A}^{(l)} \bm{D}^{-\frac{1}{2}}) \bm{X} \bm{W}^{(l)} + \bm{W}_{b}^{(l)} 
\end{equation}
where $\bm{I} \in \mathbb{R}^{N \times N}$, $\bm{D}$ is the degree matrix.

The mix-up operation originally serves as a data augmentation method to better constrain the border of the feature space, and we apply the idea of 'mix-up' here to balance the composition of two kinds of correlations, and constrain the border of Adjacent Matrix. 

Besides, different from traditional mix-up which samples weighting coefficients from beta distribution, we compute the weights by self-attention module, which dynamically assigns higher weights to the high intra-similarity between long-distance nodes, and further improves mining the semantic correlations. As the following equation suggests, LM$^3$ mixes up the explicit graph $\bm{A}^{(0)}$ with learned implicit graph $\bm{A}^{(l)}$ in a convex dynamic manner.
\begin{equation} \label{equ:mix-up}
\small
    \bm{M}^{(m)}(\bm{X}) =  f_{a}(\tilde{\bm{A}}^{(0)}(\bm{X}); \bm{\theta}_{a}) + f_{a}(\tilde{\bm{A}}^{(l)}(\bm{X}); \bm{\theta}_{a})
\end{equation}
where $f_a$ denotes self-attention network, $\bm{\theta}_{a}$ and $\bm{\theta}_{b}$ are the parameters of attention network. 
By utilizing  LM$^3$ module, the dynamic convex combination of explicit graph and learned implicit graph, we carefully capture the two kinds of time-series correlation. And the experiments can demonstrate our assumption in Section \ref{ablation}.

\vspace{-0.1cm}
\subsection{Spatial Temporal Recurrent Convolution Network} \label{section:spa-tem}
As illustrated in Figure\ref{fig:model}, we first take the learnable nodes embedding $\bm{E}$ as inputs Eq.\ref{equ:gumbel} and output the probability graph $\bm{A}^{(l)}$. 
Then, we apply the Chebyshev polynomial expansion form of graph operations for both $\bm{A}^{(l)}$ and $\bm{A}^{(0)}$ and put the encoding graph into LM$^3$ module Eq.\ref{equ:mix-up} to get $\bm{M}^{(m)}$. 
In order to further understand the spatial and temporal inter-dependencies between traffic series, our method takes $\bm{X}_{0:T}$ and $\bm{M}^{(m)}$ as input, then put it to Spatial Temporal Recurrent Graph Convolution Module (STRGC) which contains a graph convolution network and a Gate Recurrent Unit (GRU) network to updates hidden states $\bm{h}_{t}$ collectively as the equations \ref{equ:gru}. 
It is worth noting that many sequence and graph architectures can be applied here, we opt for a simple but effective one to model the dynamic and comprehensive spatial-temporal correlations, and finally outputs time series predictions $\bm{X}_{T+1:T+\tau}$.
\begin{equation}\label{equ:gru}
\small
\begin{aligned}
&\bm{z}_{t} =\sigma(\bm{M}^{(m)}([\bm{X}_{0:t}, \bm{h}_{t-1}]))  \\
&\bm{r}_{t} =\sigma(\bm{M}^{(m)}([\bm{X}_{0:t}, \bm{h}_{t-1}]))  \\
&\bm{\hat{h}}_t = tanh(\bm{M}^{(m)}([\bm{r}_t\odot \bm{h}_{t-1},\bm{X}_{0:t}])) \\
&\bm{h}_{t} =\bm{z}_t \odot \bm{h}_{t-1} + (1 - \bm{z}_t) \odot \bm{\hat{h}}_t\\
\end{aligned}
\end{equation}
where the embedded graph information is sent into reset gate $\bm{r}_{t}$ and update gate $\bm{z}_{t}$, and $\sigma$ refers to sigmoid function.
\begin{equation} \label{equ:loss}
\small
    \mathcal{L}(\bm{W}_\theta) = \frac{1}{\tau} | {\bm{X}_{T+1:T+\tau}} - \bm{\hat{X}}_{T+1:T+\tau} | 
\end{equation}
The loss function in our method is mean absolute error (MAE) loss formulated as Eq. \ref{equ:loss}.
\begin{table*}[!t]
	\vspace{-0.cm}
	\caption{Overall prediction performance of different methods on the PeMSD4 and PeMSD8 dataset, and results in boldface are the best performance achieved. (smaller value means better performance)} 
	\centering
	\small
	\scalebox{1.}{
		\begin{tabular}{c|c||ccc||ccc} 
			\toprule[1.5pt]
			\multicolumn{1}{c|}{}& {\multirow{2}{*}{Methods}} & \multicolumn{3}{c||}{PeMSD4 Dataset} &  \multicolumn{3}{c}{PeMSD8 Dataset}  \\
			\multicolumn{1}{c|}{}& {} & MAE $\bm{\downarrow}$ & RMSE $\bm{\downarrow}$ & MAPE $\bm{\downarrow}$ & MAE $\bm{\downarrow}$ & RMSE  $\bm{\downarrow}$  & MAPE $\bm{\downarrow}$ \\ 
			\midrule[1.5pt]

			\multirow{6}{*}{\rotatebox{90}{Sequence}}   & HA & 38.03       & 59.24       & 27.88\%       & 34.86       & 52.04       & 24.07\%       \\ 
			& ARIMA & 36.84       & 55.18       & -       & 34.27       & 48.88       & -       \\
			& VAR & 24.54       & 38.61       & 17.24\%       & 19.19       & 29.81       & 13.10\%       \\
			& SVR & 24.44       & 37.76       & 17.27\%       & 20.92       & 31.23       & 14.24\%       \\
			& GRU-ED & 23.68       & 39.27       & 16.44\%       & 22.00       & 36.23       & 13.33\%   \\
			& FC-LSTM ~\cite{sutskever2014sequence}  & 23.60       & 37.11       & 16.17\%       & 21.28       & 31.88       & 13.72\%  \\
			& DSANet ~\cite{huang2019dsanet}  & 22.79       & 35.77       & 16.03\%       & 17.14       & 26.96       & 11.32\%  \\
						
			\midrule[1.5pt] 
			\multirow{9}{*}{\rotatebox{90}{Graph}} & DCRNN~\cite{DCRNN}        & 21.22       & {33.44} & 14.17\%       & {16.82} & {26.36} & {10.92\%} \\   
			& STGCN ~\cite{stgcn}       & {21.16} & 34.89       & {13.83\%} & 17.50       & 27.09       & 11.29\%       \\ 
			 & ASTGCN ~\cite{guo2019attention}       & 22.93       & 35.22       & 16.56\%       & 18.25       & 28.06       & 11.64\%       \\ 
            & STSGCN ~\cite{song2020spatial}    & 21.19      & 33.65       & 13.90\%       & 17.13       & 26.86       & 10.96\%       \\ 
            & AGCRN~\cite{agcrn}       & {19.83}       & {32.26}       & {12.97\%}       & {15.95}       & {25.22}       & {10.09\%}       \\ 
            & Z-GCNETs~\cite{Z-GCNETs}       & {19.54}       &{31.33}       & {12.87\%}       & {16.12}       & {25.74}       & {10.35\%}       \\ 
            & \textbf{Ours}         & \textbf{{19.19}}      & \textbf{{31.14}}      & \textbf{{12.69\%}}     & \textbf{{15.49}}     & \textbf{{24.80}}         & \textbf{{9.96\%}}       \\
			& Improvements      & {{+1.79\%}}       &  {{+0.61\%}}       & {{+1.40\%}}          &  {{+3.91\%}}        & {{+3.65\%}}      &  {{+3.92\%}}   \\ 
			\bottomrule[1.5pt]
		\end{tabular} 
	}
	\vspace{-0.cm}
	\label{tab:sota}
\end{table*}

\vspace{-0.2cm}
\section{Experiments}
\subsection{Datasets}
To evaluate the performance of the proposed RGSL, we conduct experiments on two public real-world traffic datasets PeMSD4 and PeMSD8 ~\cite{guo2019attention,stsgcn-aaai2020}, and a proprietary dataset RPCM. 
\\\textbf{PeMSD4}: The PeMSD4 dataset refers to the traffic flow data in the San Francisco Bay Area and contains 3848 detectors on 29 roads. The time span of this dataset is from January to February in 2018
\\\textbf{PeMSD8}: The PeMSD8 dataset contains traffic flow information collected from 1979 detectors on 8 roads on the San Bernardino area from 1/Jul/2016 - 31/Aug/2016.
\\\textbf{RPCM}: The RPCM dataset collects Remote Procedure Call(RPC) data every ten minutes provided by a world leading internet company. The RPC data is a direct reflection of the remote communications and mutual calls between distributed systems. RPCM contains time series from 113 applications deployed in different Logic Data Center(LDC).

\subsection{Baselines}
To evaluate the overall performance of our work, we compare our model RGSL with traditional methods and deep models as follows.

1) Sequence models which classically focus on capturing temporal relations among series, such as \textbf{Historical Average (HA)} which uses the average of the same historical timestamps as predictions, \textbf{Vector Auto-Regression (VAR)} ~\cite{var}, \textbf{support vector regression (SVR)}~\cite{svr}, \textbf{Auto-Regressive Integrated Moving Average
(ARIMA)} ~\cite{arima} which is a statistics method modelling time dependencies;
\textbf{GRU-ED} which is GRU-based encoder-decoder framework; \textbf{FC-LSTM}~\cite{sutskever2014sequence} uses
long short-term memory (LSTM) and encoder decoder network;\textbf{Dual Self-attention Network(DSANet)} ~\cite{huang2019dsanet} using CNN and self-attention mechanism for temporal and spatial correlations respectively;

2) Deep graph models: \textbf{Diffusion convolutional recurrent neural network(DCRNN)} ~\cite{DCRNN} which adopts a diffusion process on a directed graph to model the traffic flow, utilizes bidirectional random walks, and combines GCN with recurrent models in an
encoder-decoder manner; 
\textbf{Spatio-temporal graph convolutional network(STGCN)} ~\cite{stgcn} exploits spatial convolution and temporal convolution for forecasting; 
\textbf{Attention-based spatio-temporal graph convolutional network(ASTGCN)} ~\cite{guo2019attention} which integrates attention mechanisms based on STGCN for capturing dynamic spatial and temporal patterns; \textbf{Spatial-Temporal Synchronous Graph Convolutional Network(STSGCN)} ~\cite{stsgcn-aaai2020} which addresses heterogeneities in spatial-temporal data and designs multiple modules for different time periods; \textbf{Adaptive Graph Convolutional Recurrent Network (AGCRN)} ~\cite{agcrn} which learns node-specific and data-adaptive graph to capture fine-grained features; 
\textbf{GCNs with a time-aware zigzag topological layer (Z-GCNETs)}~\cite{Z-GCNETs} introduces the concept of zigzag persistence into time-aware GCN.

\textbf{Evaluation Metrics}: 
In this paper, we use mean absolute error (MAE), mean absolute percentage error (MAPE), and root-mean-square error (RMSE) to measure the performance. 

\subsection{Results}
\begin{table*}[!t]
	\small
	\centering
	\caption{\small{Component-wise analysis. Ablation study on explicit graph and implicit graph which are inputs of LM$^3$, as well as RGG and LM$^3$.}}
    \scalebox{1.0}{
		\begin{tabular}{c|cccc|ccc|ccc}
			\toprule[1.07pt]
			{\multirow{2}{*}{$\text{\#}$}} & Explicit & Implicit & $LM^3$ & RGG &  \multicolumn{3}{c|}{RPCM Dataset} &  \multicolumn{3}{c}{PeMSD4 Dataset} \\
			{} & Graph & Graph & Module & Module & MAE $\bm{\downarrow}$ & RMSE $\bm{\downarrow}$    & MAPE $\bm{\downarrow}$ & MAE $\bm{\downarrow}$ & RMSE  $\bm{\downarrow}$ & MAPE    $\bm{\downarrow}$ \\
			\midrule[1.07pt]
			1&\cmark & & & & $0.05 $  & $0.08$ &$60.13\%$& $25.13$& $38.82$& $17.62\%$ \\
			2& & \cmark & & & $ 0.04 $  &$0.06$ &$32.86\%$& $19.84$& $32.35$& $13.01\%$  \\
			3&\cmark & \cmark & &  & $ 0.03$  &$0.06$ &$30.54\%$& $19.47$& $31.85$& $12.85\%$   \\
			4&\cmark & \cmark & \cmark &  & $0.03$  &$0.06$ &$27.86\%$& $19.37$& $31.50$& $12.82\%$   \\
			5&\cmark & \cmark & \cmark & \cmark & $\textbf{0.03}$  &$\textbf{0.05}$ &$\textbf{24.77}\%$ & $\textbf{19.19}$& $\textbf{31.14}$& $\textbf{12.69}\%$ \\
			\bottomrule[1.07pt]
		\end{tabular}
    \vspace{-0.1cm}
	}\label{tab:ablation}
	\normalsize
\end{table*}

\begin{figure}[t]
\centering
\includegraphics[width=1.\linewidth]{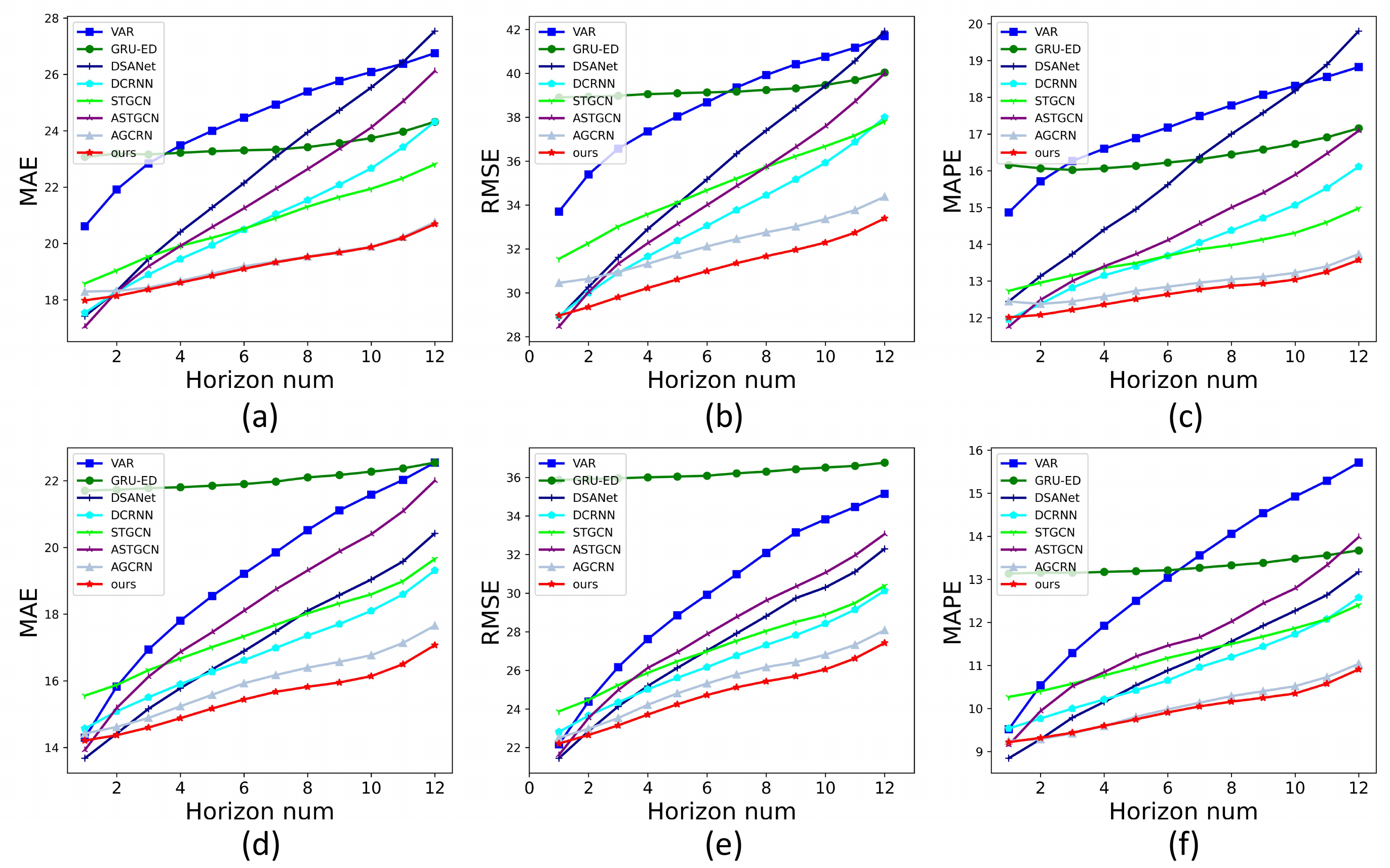}  
\caption{Evaluation metrics at each horizon in prediction period on PeMSD4 (a,b,c) and PeMSD8 (d,e,f) dataset}
\label{fig:horizons8}
\vspace{-0.3cm}
\end{figure}

\begin{figure}[t]
\centering
\vspace{-0.1cm}
\includegraphics[width=1.\linewidth]{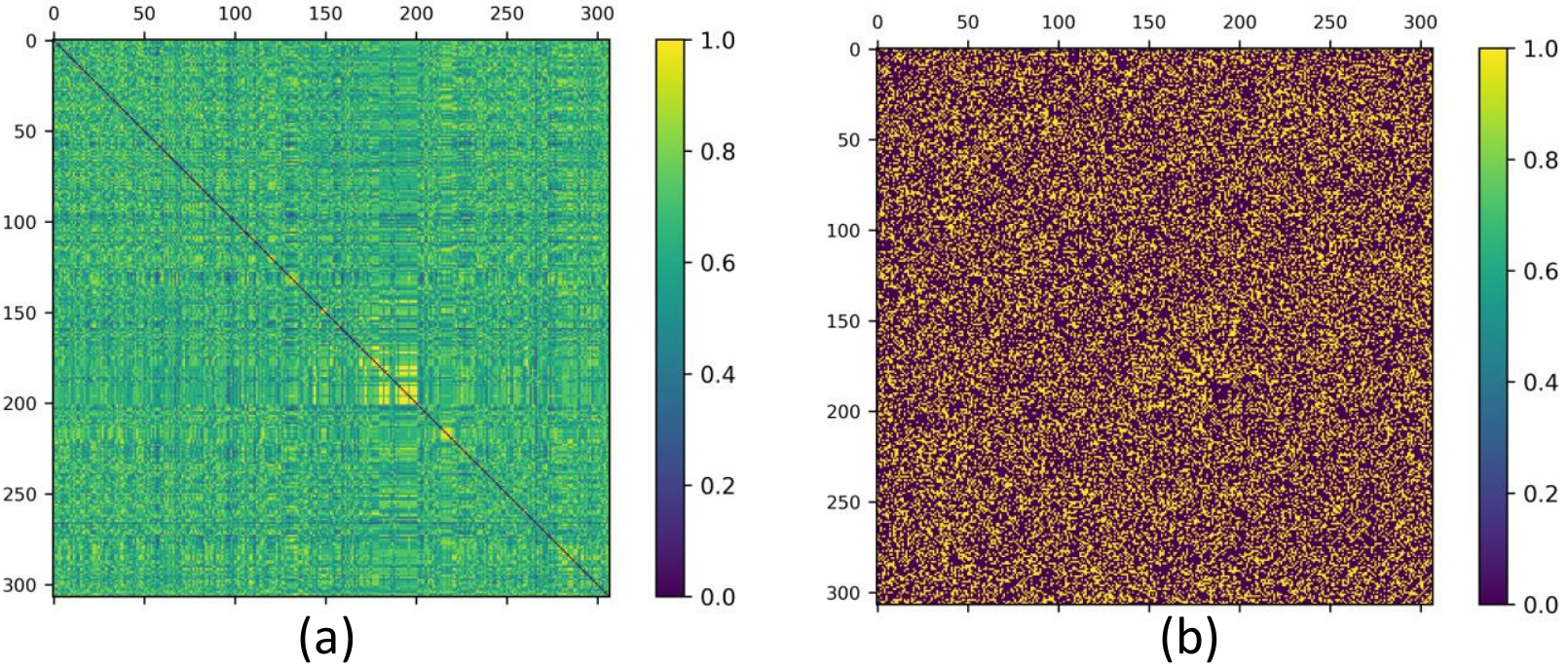}  
\caption{RGG effect on adjacent matrix; (a) are adjacent matrix after Softmax; (b)are adjacent matrix after GumbleSoftmax;}
\label{fig:adjsparce}
\vspace{-0.1cm}
\end{figure}

\begin{figure}[h]
\centering
\includegraphics[width=1.\linewidth]{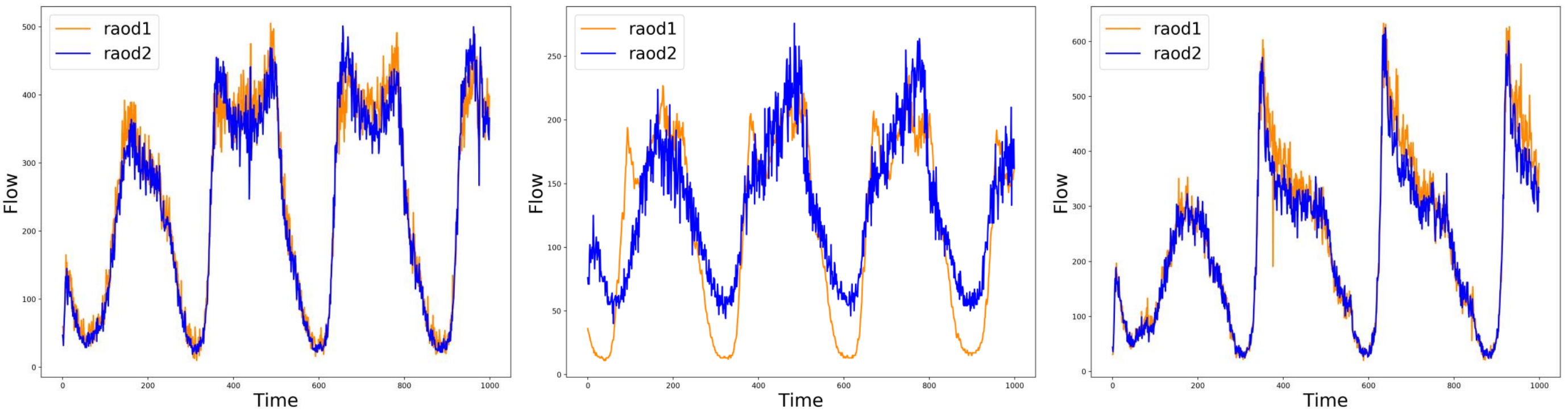}  
\caption{RGSL is able to excavate implicit relationships (road 1 and road 2). Road 1 and road2 share similar peaks and valleys, trends and timestamps.}
\label{fig:vis}
\vspace{-0.1cm}
\end{figure}

The overall experimental results, including our RGSL and other baselines are shown in Table \ref{tab:sota}.
We summarize the results in the following three aspects: 1) GCN-based deep methods outperform other methods, which is reasonable since graph pays more attention to spatial correlations; 2) Our method further improves traffic forecasting methods with a significant margin at all metrics. Our RGSL outperforms AGCRN by 3.22\% MAE on PeMSD4 and around 2.88\% MAE on PeMSD8 mainly result from 2 reasons. The first is that AGCRN totally ignores the explicit graph knowledge. Secondly, AGCRN learns dense and somehow redundant adjacent matrix. We demonstrate the above explanations after add on our module, as shown in table \ref{tab:Model agnostic}. Z-GCNETs performs better than AGCRN (relatively 1\% in MAE and MAPE, but inferior about 1\% in RMSE). 
And one of the differences is Z-GCNETs utilizes explicit graph information, demonstrating the significance of prior knowledge. Z-GCNETs processes raw data in part, regards each part as a channel and stacks channel-wise layers, which can capture more detailed features, but it also has graph redundancy problem. 3) The proposed method significantly outperforms other methods by 0.61\% - 3.92\% relative improvements corroborates that regularization plays an important role in graph networks, and carefully collaborating prior knowledge and learned correlations is promising and worthwhile. Experiments results in Figure\ref{fig:vis} shows that road 1 and road 2 are not spatially connected and thus not existed in prior explicit graph, but RGSL captures their similarity and is able to model this similarity by learning $A^{(l)}$. 
Besides, in Figure\ref{fig:horizons8}our model also shows superior for all horizons and deteriorate slower than other GCN-based models.

\vspace{-0.3cm}
\subsection{Ablation Study} \label{ablation}
Table\ref{tab:ablation} illustrates the effect of removing key components of our method on both PeMSD4 and RPCM dataset. There are two main modules LM$^3$ and RGG, and LM$^3$ takes prior graph and learnable graph as input. 

Table\ref{tab:ablation} shows four ablations that removing RGG module results in degradation(relative 1\%) due to the lack of regularization. Besides, removing RGG also makes adjacent matrix less explainable, as in Figure\ref{fig:adjsparce}(b), the learned adjacent matrix is filled with small enties. \ref{fig:adjsparce}(b) and \ref{fig:visual}(c) are adjacent matrix with LM$^3$ module, it is sparse and accord with our expectation. 

Table\ref{tab:ablation} experiment 3,2,1 and Figure \ref{fig:visual} shows the effect of LM$^3$ module, while experiment 1 and 2 respectively takes prior knowledge or learned knowledge as the only graph input. The substantial decrease in performance demonstrates the followings: 1)only the explicit graph restricts the sum of learnable parameters and lacks flexibility; 2) learnable-only graph performs far better than priori graph, yet miss plenty of useful knowledge. As shown in \ref{fig:visual}(a), compared to (c), it lacks some key edges(in red boxes); 3) utilizing both prior and learnable knowledge by simply add them up shows small gain in performances, but is inferior to our carefully designed convex dynamic pattern, since the fixed parameters lacks the ability to weight the importance of the above two matrix according to their distributions. 
Overall, our modules shows great potential to boost the prediction performance as well as can be easily employed in different graph neural networks.

\begin{table}[t]  
	\centering
	\small
	\caption{Model agnostic results on PeMSD4 dataset.}
	\label{tab:agnostic}
		\scalebox{1.0}{
			\begin{tabular}{lccc}
				\toprule[1.5pt]
				Methods  & MAE $\bm{\downarrow}$ & RMSE  $\bm{\downarrow}$ & MAPE    $\bm{\downarrow}$ 
				\\ \midrule[1.5pt]
				STGCN~\cite{stgcn} &$21.16$ & $34.89$ & $13.83$ \\
				\textbf{+Ours} & \textbf{20.21} & \textbf{33.21} & \textbf{13.22}  \\
				\midrule
				AGCRN~\cite{agcrn} & $19.83$ & $32.26$ & $12.97$\\			
				\textbf{+Ours} & \textbf{19.19} & \textbf{31.14} & \textbf{12.69} \\
				\bottomrule[1.5pt]
		\end{tabular}} \label{tab:Model agnostic}
		\vspace{-0.2cm}
\end{table}

\textbf{Model Agnostic Analysis}: Besides, RGSL can be generalized to different GNNs. As Tab \ref{tab:agnostic} shows, RGSL added on STGCN also significantly increases the metrics (relative improvements +4.49\% MAE, +4.82\% RMSE, +4.41\%MAPE ). Both the LM$^3$ and RGG modules do not have excrescent requirements for graph or graph-related tasks, and can be easily applied and extended to related tasks.
\section{Related Work}

Multi-variates time-series forecasting has been widely studied, and there are exhaustive surveys and thesis which can be 
referred to~\cite{yin2020comprehensive,thesis}. Here we focus more on the recent advances of deep neural network. 
The deep learning methods treat spatio-temporal relations mainly utilizing RNN
based~\cite{guo2020optimized,pan2020spatio}, GNN based ~\cite{stgcn,guo2019attention,song2020spatial,agcrn,Z-GCNETs}, and CNN based~\cite{wu2020gcnsurvey} networks, where CNN lacks generalization since the irregularity of input. In detailed RNN based methods, \textit{attention mechanism}~\cite{jin2020deep,li2020forecaster,zhang2020knowledge} is frequently applied to model temporal correlation, while the simpler FNN-based approach is used in~\cite{wei2019dual,stsgcn-aaai2020,cao2020spectral,chen2020tssrgcn,zhang2020spatial} can also show competitive results. However, these RNN, attention or FNN mechanism shows deficiency in performance because of their ability to corelates and interacts spatial and temporal information. Apart from capturing temporal dependency with NN, other techniques that have also been combined with GNNs include autoregression~\cite{lee2019demand}, and Kalman filters.
\\ In GNN based methods, there are multiple ways to construct adjacent matrix, which can be split into Connection Matrix~\cite{stsgcn-aaai2020}, Distance Matrix~\cite{DCRNN}, Similarity Matrix~\cite{lv2020temporal} and Dynamic Matrix. 
However, these methods discard the correlation between prior knowledge and learned re-allocate embeddings. ASTGCN~\cite{guo2019attention} utilizes spatial-temporal attention and spatial-temporal convolution to model adjacent spatial-temporal relationship, and the input temporal information is from three different granularity. AGCRN~\cite{agcrn} captures fine-grained spatial and temporal correlations in traffic series automatically based on the two modules and recurrent networks, but lacks interpretability in visualization. ST-Norm~\cite{stnorm}  propose temporal and spatial normalization and separately refine the high-frequency component and the local component underlying the raw data. 
\vspace{-0.2cm}
\section{Conclusion}
In this paper, we propose a novel model RGSL which contains a laplacian matrix mix-up module to automatically discover implicit time-series pattern and fuse both the explicit graph and implicit graph in a dynamic convex fashion. 
Furthermore, we argue that graph regularization which improves graph interpretability also helps to capture the complicated correlation more efficiently.  
Thus we propose a RGG module which is easy to add-on and has similar property to Dropout. 
Experiments on three real-world flow datasets show our model RGSL enhances the state-of-the-art performances by a large margin. 
Since the proposed LM$^3$ module and RGG module are general and do not have additional constraints on task, future works and software projects can be apply these two to various graph-based tasks which are not confined to time series forecasting.

\section*{Acknowledgments}
Yan Huang and Liang Wang were jointly supported by National Key Research and Development Program of China Grant No. 2018AAA0100400, National Natural Science Foundation of China (61721004, U1803261, and 61976132), Beijing Nova Program (Z201100006820079), Key Research Program of Frontier Sciences CAS Grant No. ZDBS-LY-JSC032, CAS-AIR.
	

%



\small
\newpage
\bibliographystyle{named}
\bibliography{ref_short}

\end{document}